\documentclass[runningheads]{llncs}
\usepackage[T1]{fontenc}
\usepackage{graphicx}
\usepackage{amssymb}
\usepackage{amsmath}
\usepackage{algorithm}
\usepackage{algorithmic}
\usepackage{graphicx}
\usepackage{textcomp}
\usepackage{xcolor}
\usepackage{verbatim}
\usepackage{array}
\usepackage{fancyvrb}
\usepackage{hyperref}
\usepackage{url}
\usepackage{graphicx}
\usepackage{booktabs}
\usepackage{siunitx} 
\usepackage{adjustbox}
\usepackage{tcolorbox}
\usepackage{mdframed}
\usepackage{xcolor}
\usepackage{multirow}
\usepackage{colortbl}
\usepackage{booktabs}
\usepackage{ulem}
\usepackage{tabularx}
\usepackage{soul}
\usepackage{seqsplit}
\usepackage{makecell}
\usepackage{threeparttable}
\usepackage[misc]{ifsym}

\begin{document}
\title{AgentFactory: Towards Automated Agentic System Design and Optimization}
\titlerunning{AgentFactory: Towards Automated Agentic System Design and Optimization}
%
\author{
    Enci Zhang \and Haofeng Wang \and Yuesheng Zhu \and Xiaole Cui \and \\Guibo Luo\textsuperscript{\Letter}
}
\authorrunning{Enci et al.}

\institute{Peking University Shenzhen Graduate School , Shenzhen, China \\
\email{eczhang@stu.pku.edu.cn, hfwang@stu.pku.edu.cn, zhuys@pku.edu.cn, cuixl@pkusz.edu.cn, luogb@pku.edu.cn}}
%
%
\maketitle              
\begin{abstract}
Large Language Models (LLMs) have demonstrated remarkable capabilities as powerful components in agentic systems, enabling sophisticated reasoning and complex task execution.
However, 
current approaches to manually designing and optimizing agentic systems heavily rely on manual effort, limiting their adaptability and scalability. 
Recent work has explored the automated optimization of workflow designs. However, these approaches often overlook the crucial role of model capabilities and focus on single performance metrics, failing to address real-world deployment constraints.
In this paper, we present AgentFactory, a framework that jointly optimizes both foundation models and workflow structures in agentic systems while considering multiple objectives including performance, cost, and efficiency.
AgentFactory leverages advanced LLMs as optimizers to navigate the vast search space of possible configurations, employing a three-stage optimization pipeline to automatically discover effective combinations of fine-tuned models and optimized workflows. 
Through an iterative optimization process, our framework systematically explores and evaluates different agentic system designs, adapting to task-specific requirements while maintaining operational efficiency. 
We evaluate AgentFactory across eight benchmarks spanning five domains, including general reasoning, coding, mathematics, medicine, and finance. 
Our experiments demonstrate that AgentFactory consistently outperforms both manually designed methods and existing automated approaches, achieving an average improvement of 9.1\% across all benchmarks, with particularly significant gains in domain-specific tasks (19.6\% on MedQA and 18.7\% on FinEval). 
These results establish AgentFactory as a promising approach for developing more capable and efficient agentic systems through automated optimization.
\keywords{Large language model  \and Agent framework \and LLM fine-tuning \and Automated agentic optimization}
\end{abstract}

\section{Introduction}
\label{Introduction}
Large Language Models (LLMs)  such as GPT-4o, Claude-3.5-sonnet, and DeepSeek V3 have demonstrated remarkable capabilities as powerful components in agentic systems, enabling sophisticated reasoning and complex task execution. 
However, the effective deployment of LLM-based agentic system heavily depends on manual design and tuning.
For instance, frameworks like ChatDev~\cite{intro_chatdev_2023} and MetaGPT~\cite{related_work:metagpt_2023} predefine various roles, workflow structures, and their corresponding responsibilities, manually assigning distinct profiles to each agent to facilitate collaboration.
This approach, while functional, is labor-intensive and relies heavily on empirical experience, often leading to sub-optimal solutions and making it challenging to explore the vast solution space and discover optimal configurations.
The history of artificial intelligence reveals a consistent pattern where learned solutions supersede manual engineering—from features in computer vision to model architectures themselves via Neural Architecture Search (NAS)~\cite{intro_NAS_2023}. This historical precedent strongly suggests that automating the design of agentic systems could yield more effective and scalable solutions than current manual approaches.

Given the impressive capabilities of Large Language Models in understanding complex instructions and generating creative solutions, recent efforts have begun exploring their potential as optimizers for automated agentic system optimization, primarily focusing on prompt engineering and workflow design. 
Initial approaches like PromptBreeder~\cite{related_work:fernando2023promptbreeder_2023} and TextGrad~\cite{related_work:yuksekgonul2024textgrad_2024} have demonstrated success in automated prompt optimization. 
However, optimizing prompts alone may not be sufficient for complex tasks that require coordinated interactions between multiple components. Recognizing this limitation, more comprehensive frameworks like ADAS~\cite{ADAS_2024} and AFlow~\cite{aflow_2024} have extended automation to workflow design and optimization.
However, these approaches face two significant limitations. 
First, they concentrate solely on workflow optimization while neglecting model-level enhancements, which particularly limits their effectiveness in specialized domains where workflow optimization alone cannot bridge the domain adaptation gap. 
Second, these methods typically optimize for single metrics like accuracy or F1 scores, overlooking crucial real-world considerations such as inference costs and response latency.

To address these limitations, we present AgentFactory, a framework designed for automated agentic system generation and optimization. 
Given a task specification, AgentFactory first generates a basic agentic system and then iteratively optimizes it until all constraints are satisfied. 
The optimization process operates in the joint space of model parameters and workflow representations, subject to multiple constraints. Unlike previous approaches that only focus on workflow optimization, AgentFactory expands the optimization search space by incorporating foundation model fine-tuning, enabling more comprehensive system optimization. 
By incorporating fine-tuning into the optimization process and leveraging powerful LLMs as optimizers, our framework simultaneously optimizes multiple objectives including performance, cost, and efficiency.
Our extensive experiments across 5 domains and 8 benchmarks show an average performance improvement of 9.1\% over existing methods, with particularly significant improvements in specialized domains, achieving 19.6\% and 18.7\% improvements on MedQA~\cite{dataset:medqa_2021} and FinEval~\cite{dataset:fineval_2023} benchmarks respectively, while maintaining lower inference costs.

Overall, our contributions can be summarized as follows: 
\begin{itemize}
    \item We propose AgentFactory, a comprehensive framework for automated agentic system design and optimization.
    \item We use optimization theory to formalize the problem of agentic system optimization and develop algorithms to solve it. We further introduce model fine-tuning into the automated agent design process and pioneer multi-objective optimization considering accuracy, cost, and latency.
    \item We conduct extensive experiments across diverse domains, demonstrating superior performance compared to both manual designs and state-of-the-art automated approaches.
\end{itemize}
\section{Related Work}
\subsection{LLM-powered Agentic Systems}
Agent systems have evolved through distinct phases, from early symbolic architectures emphasizing expressiveness but lacking flexibility, to reinforcement learning approaches that achieved task-specific success but struggled with long-term planning. These limitations highlighted the need for more sophisticated architectures.

The integration of LLMs has brought unprecedented opportunities to this field. With their extensive general knowledge and logical reasoning capabilities, LLMs can serve as flexible components in complex problem-solving systems - functioning similarly to humans' automatic thinking (System 1), while the overall agent architecture enables deliberate reasoning (System 2)~\cite{related_work:thinking_fast_and_slow_2011}. This has sparked numerous innovations in agent design: Tree of Thoughts (ToT)~\cite{tree_of_thought_2024} and Graph of Thoughts (GoT)~\cite{related_work:graph_of_thought_2024} implement structured reasoning frameworks, HuggingGPT~\cite{hugginggpt_2024}  demonstrate effective tool integration, while Generative Agents~\cite{generative_agents_2023} and Reflexion~\cite{reflexion_2024} explore memory mechanisms for experience accumulation.

Recent research has further expanded these capabilities through multi-agent frameworks and system optimizations. Works like Camel~\cite{related_work:camel_2023}, and MetaGPT~\cite{related_work:metagpt_2023} leverage inter-agent discussions and role-playing to enhance reasoning and coordination. Meanwhile, platforms like AutoGPT~\cite{related_work:autogpt_2023} focus on autonomous task decomposition and execution. While these advances show promise, significant challenges remain in standardizing architectures and ensuring reliable performance across diverse tasks.

\subsection{Automated Agentic System Optimization}
The automation of agentic system design has emerged as a critical research direction, with existing approaches broadly falling into three categories: prompt optimization, hyperparameter tuning, and workflow structure optimization. 
The first category, represented by recent works~\cite{related_work:fernando2023promptbreeder_2023}~\cite{related_work:yuksekgonul2024textgrad_2024}~\cite{related_work:khattab2024dspy_2024}~\cite{related_work:yang2024buffer_2024}, explores techniques to automatically refine prompts while maintaining the original workflow structure. 
The second category~\cite{related_work:saad2024archon_2024} focuses on optimizing predefined parameters. Despite their contributions to performance enhancement, both approaches face challenges in adapting to diverse tasks and still necessitate substantial task-specific manual intervention.

The third category focuses on comprehensive workflow structure optimization, attracting significant research attention~\cite{related_work:zhu2024large_2024}~\cite{gptswarm}~\cite{ADAS_2024}~\cite{aflow_2024}. Among these efforts, GPTSwarm~\cite{gptswarm} pioneered the application of graph-based representations combined with reinforcement learning, though its effectiveness is hampered by limitations in handling complex conditional logic. ADAS made progress by adopting code-based workflow representations, but its potential is not fully realized due to constraints in its search methodology. Building upon these foundations, AFlow introduced a more sophisticated approach through its named node architecture and specialized MCTS-based optimization strategy, enabling more nuanced workflow representations and efficient search capabilities.

However, a common limitation across all these approaches is their exclusive focus on workflow optimization while keeping the underlying models fixed. This constraint significantly limits the potential performance improvements, as model capabilities play a crucial role in agent behavior. Our work addresses this limitation by introducing AgentFactory, a framework that jointly optimizes both foundation models and workflow designs. Through an iterative search over the joint space of model parameters and workflow representations, our approach demonstrates superior performance compared to methods that optimize workflow alone, establishing a new paradigm for comprehensive agent optimization.

\section{Method}
\begin{algorithm}
\caption{LLM-guided Multi-Objective Optimization for Agentic Systems}
\label{alg:af}
\begin{algorithmic}[1]
\REQUIRE Task $T$, evaluation functions $\mathbf{G}$, model space $\mathcal{M}$, LLM optimizer $\mathcal{L}_{\theta}$, meta-prompt $I_{meta}$, maximum steps $K_{max}$, target scores $\mathbf{G}_{target}$ and scalarization function $\mathcal{F}$
\ENSURE Optimal Agentic System $s^{*}$
\STATE Initialize optimization trajectory $\mathcal{O}_0 \leftarrow \emptyset$ 
\STATE Generate initial system $s_0 \leftarrow \mathcal{L}_{\theta}(\mathcal{O}_0, \mathcal{M}, I_{meta})$
\STATE Evaluate $\mathbf{G}(s_0,T)$
\STATE $\mathcal{O}_0 \leftarrow \{(s_0,\mathbf{G}(s_0,T))\}$
\FOR{$k \leftarrow 1$ to $K_{max}$} 
    \STATE $s_k \leftarrow \mathcal{L}_{\theta}(\mathcal{O}_{k-1}, \mathcal{M}, I_{meta})$ 
    \STATE Evaluate $\mathbf{G}(s_k,T)$
    \STATE  $\mathcal{O}_k \leftarrow \mathcal{O}_{k-1} \cup \{(s_k,\mathbf{G}(s_k,T))\}$ 
    \IF{$\mathbf{G}(s_k,T) \succeq \mathbf{G}_{target}$}
        \RETURN $s_k$  
    \ENDIF
\ENDFOR 
\STATE $s^{*} = \arg\max_{s \in \mathcal{O}_k} \mathcal{F}(\mathbf{G}(s, T)),$ 
\RETURN $s^*$
\end{algorithmic}
\end{algorithm}

\subsection{Problem Formulation}
We formalize the automated agentic system optimization problem as follows:  Given a task $T$ and a set of evaluation functions $\mathbf{G} = \{G_1, G_2, ..., G_p\}$ corresponding to different objectives (such as accuracy, cost, and latency), AgentFactory iteratively generates a sequence of agentic systems $\{s_0, s_1, \dots ,s_n\} \in \mathcal{S}$, which can be represented as:
\begin{equation}
    \mathcal{S} = \{(m, w)|m \in \mathcal{M}, w \in \mathcal{W}\},
\end{equation}
where $\mathcal{M}$ denotes the space of LLM-based models, formulated as:
\begin{equation}
    \mathcal{M} = \{\phi(b, d, h) | b\in \mathcal{B},d \in \mathcal{D}, h\in \mathcal{H}\},
\end{equation}
where $\mathcal{B}$ denotes the available foundation models, $\mathcal{D}$ represents the instruction fine-tuning datasets, and $\mathcal{H}$ encompasses the hyperparameters during fine-tuning, such as batch size and learning rate. $\phi$ denotes the tuning method, which could be LoRA~\cite{lora_2021}, QLoRA~\cite{qlora_2024} or full-parameter fine-tuning. For third-party models where local fine-tuning is not feasible, we set $d = \emptyset$ and $h = \emptyset$ to indicate the use of the original model.

The workflow space $\mathcal{W}$ can be represented in various forms, such as graph-based structures like GPTSwarm, neural architectures like DyLAN~\cite{method_dyline_2024}, or executable code like ADAS.

At the $k$-th evolution step, let the optimization trajectory $\mathcal{O}_{k-1}$ be defined as a sequence of solution-evaluation pairs:
\begin{equation}
    \mathcal{O}_{k-1} = \{(s_{i},\mathbf{G}(s_{i},T))\}^{k-1}_{i=0},
\end{equation}
where $s_i \in \mathcal{S}$ represents the $i$-th Agentic System and 
\begin{equation}
    \mathbf{G}(s_i,T) = \{G_1(s_i,T), G_2(s_i,T), ..., G_p(s_i,T)\}
\end{equation}
denotes a vector of evaluation metrics on task $T$.

From an optimization theory perspective, this formulation represents a complex multi-objective optimization problem over non-differentiable and partially observable spaces $\mathcal{M}$ and $\mathcal{W}$, traditional optimization methods like gradient descent or evolutionary algorithms may not be directly applicable. One feasible approach is to leverage LLMs as optimizers~\cite{llm_as_opt_2024}.
Specifically, a LLM optimizer $\mathcal{L}_{\theta}$ with parameters $\theta$ can be employed to generate the next solution using meta-prompt $I_{meta}$:
\begin{equation}
s_{k} = \mathcal{L}_\theta(\mathcal{O}_{{k-1}}, \mathcal{M}, I_{meta}).
\end{equation}

The optimization process terminates under either of the following conditions:
\begin{equation}
    \mathbf{G}(s_k, T) \succeq \mathbf{G}_{target} \text{ or } k = K_{max},
\end{equation}
where $\mathbf{G}_{target}$ represents the vector of target thresholds for each evaluation metric, $\succeq$ denotes component-wise comparison (i.e., each metric meets or exceeds its corresponding target), and $K_{max}$ is the maximum number of optimization steps.
The final objective is to find the Pareto-optimal agentic system $s^{*}$ that represents the best trade-off among competing objectives:
\begin{equation}
    s^{*} = \arg\max_{s \in \mathcal{O}_k} \mathcal{F}(\mathbf{G}(s, T)),
\end{equation}
where $\mathcal{O}_{K}$ represents the final optimization trajectory and $\mathcal{F}$ is a scalarization function that aggregates multiple objectives according to user preferences.
The end-to-end algorithm is formally presented in Algorithm \ref{alg:af}.

\begin{figure*}[t]
    \centering
    \includegraphics[width=\textwidth]{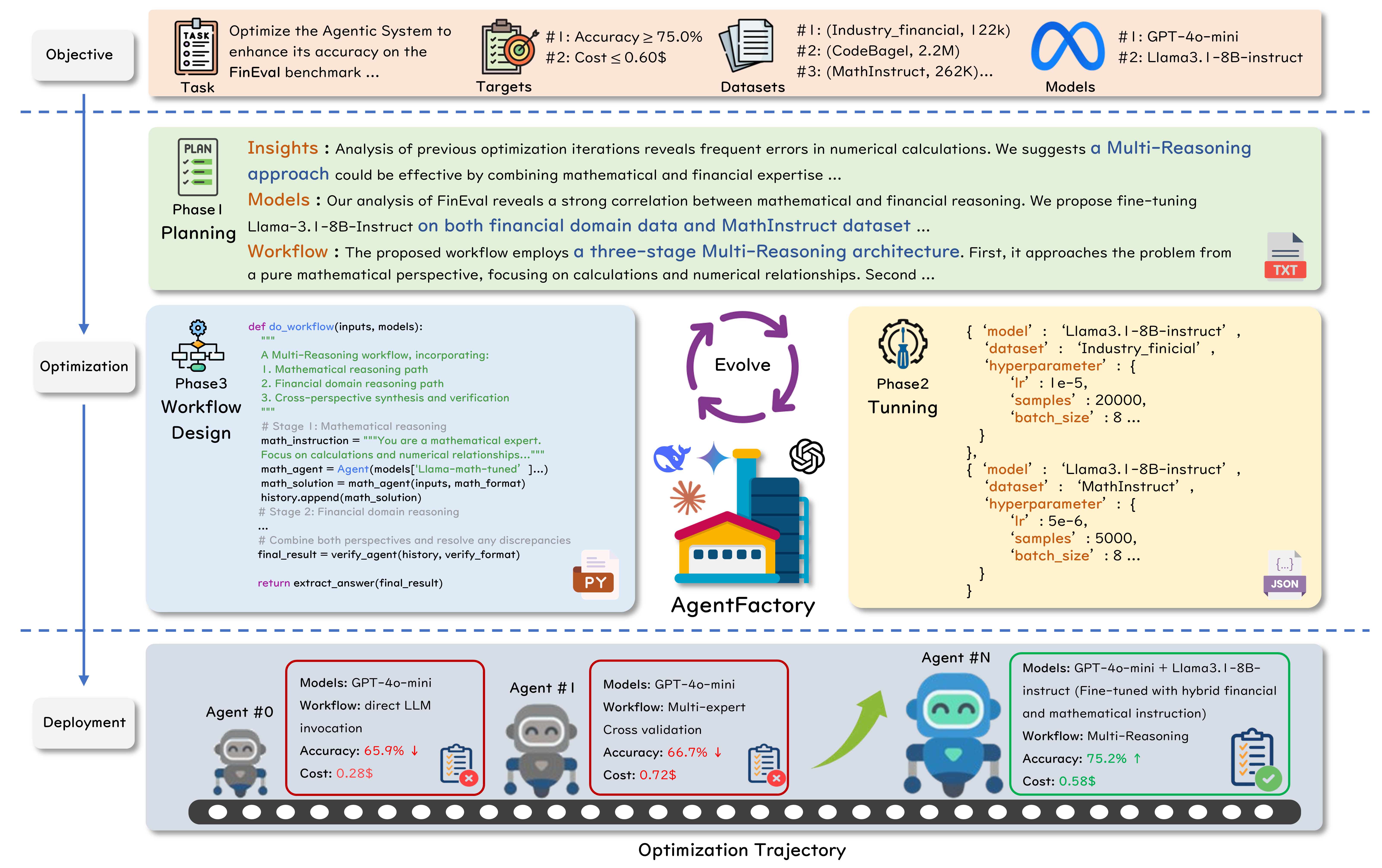} 
    \caption{The AgentFactory framework. The process begins with defining Objectives (specifying performance targets and constraints), proceeds through the Optimization stage, and concludes with Deployment of the optimized agentic
     system. During the central Optimization phase, AgentFactory employs a three-stage pipeline (planning, tuning, and workflow design) to iteratively optimize both foundation models and workflow structures until meeting the specified requirements, at which point the system becomes ready for deployment on the target task.}
    \label{fig:af_arch}
\end{figure*}

\subsection{AgentFactory Overview}
We propose AgentFactory, a comprehensive framework for automated agentic system design and optimization. As illustrated in Fig.\ref{fig:af_arch}, current approaches to agentic system design either rely on manual design processes or restrict their focus to workflow optimization alone. 
AgentFactory addresses these limitations by introducing a three-stage Optimization Pipeline (planning, tuning, and workflow design) that simultaneously optimizes both model parameters and workflow structures in the joint search space, enabling more comprehensive system optimization.

For a given task and associated performance targets, AgentFactory formalizes the optimization process as a multi-objective search over the joint space of model parameters and workflow structures. The optimization trajectory shown in the bottom right of Fig.\ref{fig:af_arch} demonstrates how the system progressively converges toward solutions that satisfy all predetermined performance target.

At the core of AgentFactory is a sophisticated three-stage optimization pipeline that systematically decomposes the complex search problem. This structured approach enables efficient navigation of the optimization landscape while ensuring thorough exploration of the solution space. As shown in Fig.\ref{fig:af_arch}, the process progresses from objective definition through iterative optimization and ultimately to deployment once all performance, cost, and efficiency targets are met. 
\subsection{Optimization Pipeline}
From an optimization theory perspective, the joint search over model parameters and workflow structures presents a challenging multi-objective optimization problem with a vast and complex solution space. To address this complexity, AgentFactory employs a three-phase decomposition strategy that balances exploration of diverse solutions with exploitation of promising directions.

The decomposition into distinct phases—planning, tuning, and workflow design—rather than attempting simultaneous optimization of all components, is essential for managing the inherent complexity of agentic system design. This structured approach allows precise control over format-sensitive outputs while maintaining system coherence through a unified optimization framework. Furthermore, this decomposition reduces coupling between different optimization aspects, enabling focused improvements in each dimension of the search space.

\definecolor{leftbarcolor}{RGB}{51,122,183} 
\begin{figure*}[h]
\noindent
\begin{mdframed}[
    backgroundcolor=gray!10,
    leftline=true,
    linewidth=3pt,
    linecolor=leftbarcolor,
    rightline=false,
    topline=false,
    bottomline=false,
    innerleftmargin=10pt, 
    innerrightmargin=10pt,
    skipabove=-2em, 
    skipbelow=0.5em]
\small
\noindent\textbf{Main prompt for the LLM optimizer}
\medskip

\noindent \textbf{ROLE}: You are an Agentic System optimization expert specialized in improving LLM-based agent architectures and workflows. You understand the intricacies of agent coordination, task decomposition, and the dynamic adaptation of agent behaviors based on performance feedback and system requirements.

\noindent \textbf{TASK}: Your task is to optimize the Agentic System to enhance its performance on \{TASK\_NAME\}. Based on the provided optimization trajectory, propose an improved system design by selecting appropriate models and designing efficient workflows.

\medskip
\noindent \{Task Description\}

\noindent \{Model Configuration\}

\noindent \{Workflow Specification\}

\noindent \{Optimization Trajectory\}
\medskip

\noindent Based on the above information, please provide your solution in three parts:

\hspace{2em}\textbf{Insights}: Explain your design rationale and key optimization strategies.

\hspace{2em}\textbf{Model}: Specify the model selection and fine-tuning configurations.

\hspace{2em}\textbf{Workflow}: Detail the implementation of your optimized workflow design.
\end{mdframed}
\caption{The main prompt template for the LLM optimizer $\mathcal{L}_\theta$ in AgentFactory framework. The content within curly braces represents variable components, including the meta-prompt $I_{meta}$ that guides the optimization process, the optimization trajectory $\mathcal{O}_{k-1}$ containing historical solutions and their performance, and specifications of the LLM-based model space $\mathcal{M}$. }
\label{fig:prompt}
\end{figure*}

\textbf{Planning}:
The planning phase serves as the cornerstone of our optimization framework, where the LLM optimizer synthesizes comprehensive optimization strategies based on multiple information sources. 
During each iteration, the optimizer integrates the meta-prompt with task descriptions, model configurations, workflow design frameworks, and previous optimization trajectories to generate a structured optimization plan. 
This plan comprises three essential components: 
(1) Insights that articulate the strategic rationale behind the proposed optimizations, 
(2) Model specifications encompassing all model-related configurations, including foundation model selection, appropriate datasets, fine-tuning approaches, and specific hyperparameter settings , and
(3) Workflow designs outlining the systematic implementation approach. 
As shown in Figure \ref{fig:prompt}, we developed a specialized prompt template that guides the LLM optimizer in generating these structured plans. 

\textbf{Tuning}:
Based on the optimization plan, the LLM optimizer generates specific tuning configurations, encompassing dataset selection, data samples requirements, fine-tuning methodologies, and hyperparameter specifications. 
These configurations are then executed by the tuning backend to modify the foundation models. The separation of tuning from the planning phase provides enhanced control over the fine-tuning process, which is particularly crucial given the resource-intensive nature of LLM fine-tuning. 
In practice, the initial tuning configurations may require iterative refinement due to hardware constraints or computational limitations. 
Our framework accommodates this reality by enabling dynamic adjustment of tuning parameters through LLM-guided optimization, such as batch size reduction or adoption of more efficient fine-tuning approaches, ensuring robust performance within available computational resources.

\begin{figure}[!h]
    \centering
    \includegraphics[width=0.5\linewidth]{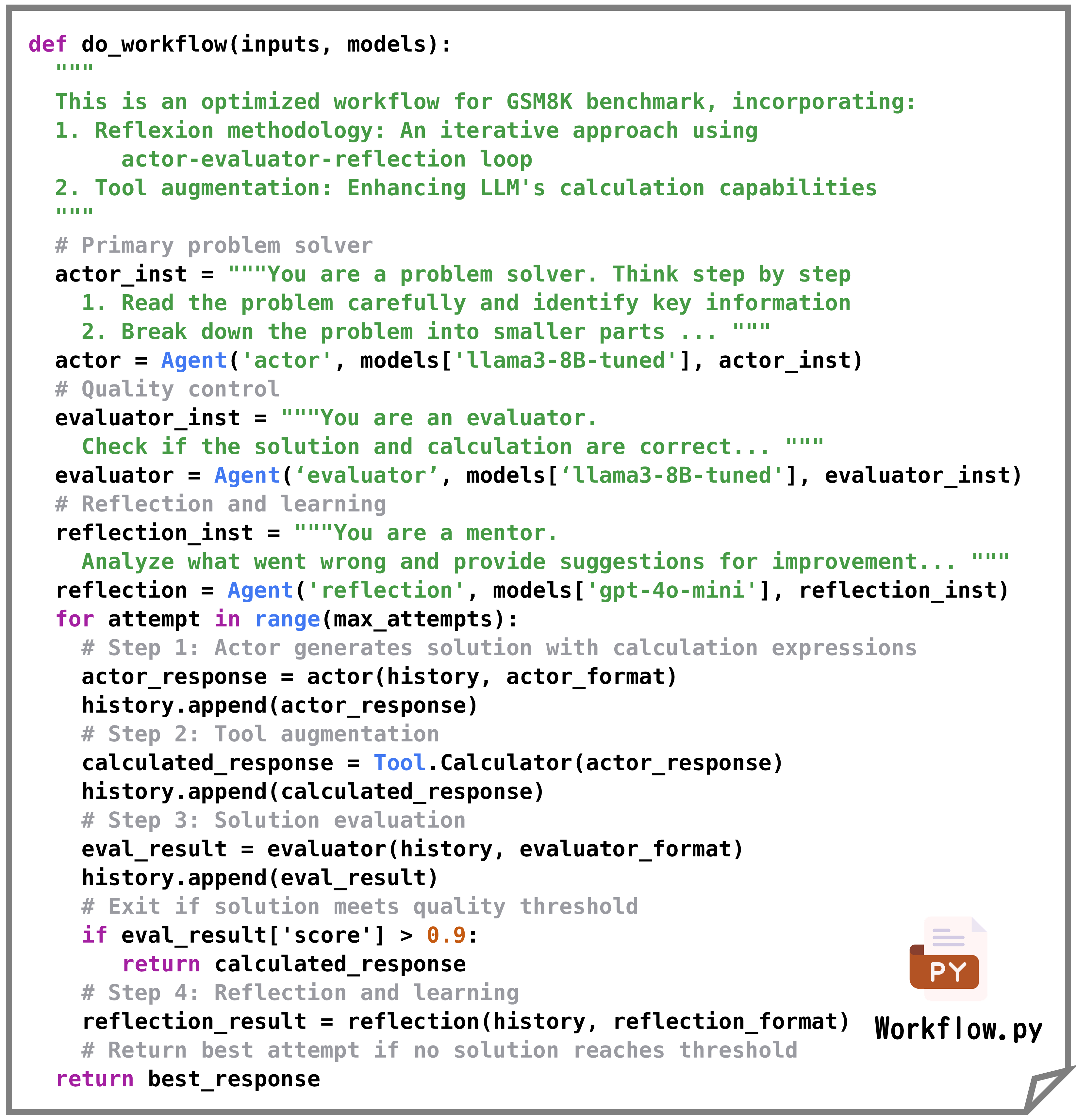} 
    \caption{An example workflow generated by AgentFactory for the GSM8K mathematical reasoning task, guided by the optimization plan. The implementation incorporates a Reflexion-based approach with specialized models and tool augmentation for numerical calculations, demonstrating the realization of optimized workflow in code space.}
    \label{fig:workflow_example}
\end{figure}

\textbf{Workflow Design}:
During the workflow design phase, the system generates executable code representations based on the optimization plan. 
Our choice of code-based workflow representation offers several key advantages over alternative approaches.
First, the Turing-completeness of programming languages enables the representation of arbitrary computational processes and control flows, providing superior expressiveness and flexibility in system design. For example, compared to graph-based approaches, code-based representations can more naturally express complex control structures and dynamic behaviors.
Second, this approach leverages the strong code generation capabilities of modern LLMs, allowing for more sophisticated and nuanced workflow implementations. 
Following ADAS's foundational work, we adopt code-based workflow representations that can capture complex agent behaviors, including prompting strategies, tool utilization, and control flow mechanisms. 
As shown in Figure \ref{fig:workflow_example}, based on our provided code templates, the LLM optimizer generates structured code that is seamlessly integrated into the agentic system. 

\section{Experiment Setup}
\subsection{Benchmarks}
We evaluated AgentFactory across five distinct domains using eight carefully selected public benchmarks. 
For general domain evaluation, we randomly selected 1,000 test instances each from MMLU~\cite{dataset:MMLU_2021} and DROP~\cite{dataset:DROP_2019}. 
For the coding domain, we utilized the complete test sets of HumanEval~\cite{dataset_humaneval_2021} and MBPP~\cite{dataset_mbpp_2021}. 
In mathematical reasoning, we employed the full GSM8K~\cite{dataset_gsm8k_2021} test set and, following~\cite{related_work:metagpt_2023}, selected 617 problems from MATH~\cite{dataset_math_2021} at difficulty level 5, spanning four representative problem types (Combinatorics \& Probability, Number Theory, Pre-algebra and Pre-calculus). 
To assess domain-specific capabilities, we included the complete test sets of MedQA~\cite{dataset:medqa_2021} and FinEval~\cite{dataset:fineval_2023} for medical and financial domains, respectively. 

\subsection{Compared Methods}
We conducted comprehensive comparisons between AgentFactory and various existing agent design approaches. 
For manually designed methods, we compared against direct LLM invocation (IO), Chain-of-Thought~\cite{compare_method:cot_2022}, Self-Consistency CoT~\cite{compare_method:cotsc_2022}with 5 samples, Reflexion with 3 rounds, and LLM Debate~\cite{compare_method:llm_debate_2023} with 3 rounds. 
For automated workflow optimization, we included two state-of-the-art baselines: ADAS and AFlow.

To ensure fair comparison, we conducted experiments across two settings with different foundation models. In the first setting, all methods utilized Llama-3.1-8B-Instruct as the foundation model. 
In the second setting, GPT-4o-mini served as the foundation model. Given AgentFactory's capability for automatic model selection and multi-objective optimization, we allowed it to dynamically leverage both Llama-3.1-8B-Instruct and GPT-4o-mini to achieve optimal balance between performance and cost efficiency.

\subsection{Implementation Details}
The performance of AgentFactory's fine-tuning process heavily depends on data quality, thus we carefully curated a diverse collection of datasets including training splits from our evaluation benchmarks, CodeBagel's~\cite{sft_data:code_bagel_2024} Python-specific segments for programming tasks, MathInstruct~\cite{sft_data:mathinstruct_2023} for mathematical reasoning, and IndustryInstruction's~\cite{sft_data:IndustryInstruction_2024} Health-Medicine and Finance-Economics subsets for domain-specific knowledge. 
These datasets are presented to AgentFactory as descriptive triples containing dataset name, size, and characteristics, enabling automated selection and configuration  based on optimization objectives. 
We employed Llama-3.1-8B-Instruct as our base model for fine-tuning, utilizing OpenRLHF~\cite{openrlhf_2024} as the backend framework. To maintain computational efficiency while ensuring stable performance, we imposed constraints of 20K samples maximum and 4-hour time limit per fine-tuning session, with experiments conducted on 4 A6000 GPUs (48GB each). 
A caching mechanism was implemented to store fine-tuned models, preventing redundant training under identical parameters. 
For the LLM optimizer component, we evaluated several models including GPT-4o\footnote{https://openai.com/} (2024-08-06), GPT-4o-mini (2024-07-18), Claude-3.5-sonnet\footnote{https://claude.ai/} (2024-10-22), Gemini Pro 1.5\footnote{https://gemini.google.com/} and DeepSeek V3\footnote{https://www.deepseek.com/}, with optimization iterations limited to 20 steps to balance exploration and computational resources.

\section{Experiment Results}

\begin{table*}[!h]
    \centering
    \caption{Performance Comparison Across Different Base Models and Methods on Multiple Benchmarks}
    \scriptsize
    \renewcommand{\arraystretch}{1.4} 
    \begin{threeparttable}
    \begin{tabularx}{\textwidth}{l|l|XXXXXXXX|p{0.6cm}p{0.6cm}}
        \toprule
        \multirow{2}{*}{\textbf{FM$^{\mathrm{a}}$}} & \multirow{2}{*}{\textbf{Method}} & \multicolumn{8}{c|}{\textbf{Benchmarks}} & \multirow{2}{*}{\textbf{Avg.}}  & \multirow{2}{*}{\textbf{Cost(\$)$^{\mathrm{d}}$}} \\
        \cline{3-10}
            & \rule{0pt}{5ex} & \textit{MMLU} & \textit{DROP} & \makecell{\textit{Human}\\\textit{Eval}} & \textit{MBPP} & \textit{GSM8K} & \textit{MATH} & \textit{MedQA} & \makecell{\textit{Fin}\\\textit{Eval}} &&\\
        \midrule
            \multirow{8}{*}{Llama$^{\mathrm{b}}$} 
                & IO        & 66.5      & 49.0      & 67.0      & 44.7      & 65.1      & 14.5      & 41.5      & 33.8      & 43.5     & \textbf{0.03} \\
                & CoT       & 69.4      & 58.2      & 72.6      & 51.4      & 82.3      & 23.5      & 45.7      & 39.6      & 51.9     & \uline{0.07} \\
                & CoT-SC    & 68.3      & 59.8      & 71.3      & 53.7      & 85.7      & 23.8      & 45.1      & 38.2      & 52.1     & 0.15 \\
                & Reflexion & 72.1      & 64.5      & 73.1      & 58.8      & 87.4      & 28.7      & \uline{47.3}      & \uline{41.2}      & 56.0     & 0.19 \\
                & Debate    & 70.3      & 62.3      & 68.9      & 48.2      & 81.2      & 22.5      & 44.8      & 37.1      & 52.3     & 0.17 \\
                & ADAS      & 73.2      & 63.8      & 73.8      & 59.9      & 88.3      & 28.4      & 46.7      & 40.3      & 56.7     &  0.36\\
                & AFlow     & \uline{72.9}& \uline{66.2} & 75.0      & 62.3      & \uline{89.4} & \uline{31.1}      & 45.8      & 39.8      & \uline{59.8}     & 0.29 \\
                & \cellcolor{gray!20}Ours 
                & \cellcolor{gray!20}\textbf{79.8}  
                & \cellcolor{gray!20}\textbf{67.4}  
                & \cellcolor{gray!20}\textbf{84.4}  
                & \cellcolor{gray!20}\textbf{70.8}  
                & \cellcolor{gray!20}\textbf{92.2}  
                & \cellcolor{gray!20}\textbf{44.6}  
                & \cellcolor{gray!20}\textbf{65.4}  
                & \cellcolor{gray!20}\textbf{58.5}  
                & \cellcolor{gray!20}\textbf{68.9}  
                & \cellcolor{gray!20}0.26  \\
        \hline
            \multirow{8}{*}{GPT$^{\mathrm{c}}$} 
                & IO        & 70.3      & 67.7      & 79.3      & 69.3      & 78.8      & 48.6      & 66.8      & 65.9     & 67.7     & \textbf{0.34} \\
                & CoT       & 74.6      & 78.9      & 81.7      & 71.6      & 91.8      & 62.7      & 74.9      & 66.2      & 74.8     & 0.72 \\
                & CoT-SC    & 75.8      & 76.3      & 81.7      & 72.8      & 91.9      & 63.5      & 73.5      & 65.8      & 74.7     & 1.43 \\
                & Reflexion & 77.9      & 78.4      & 83.5      & 76.3      & \uline{93.1}& 67.7      & 78.9    & 67.1      & 77.4    & 1.62 \\
                & Debate    & 72.3      & 75.8      & 80.0      & 72.4      & 90.5      & 59.8      & 77.3      & 65.4      & 73.7    & 1.59 \\
                & ADAS      & \uline{78.9} & 76.3      & 83.5      & 77.8      & 92.9      & 63.0      & 77.5      & \uline{67.3}      & 76.7     & 2.37 \\
                & AFlow     & 78.7 & \textbf{79.9} & \uline{86.6}   & \uline{78.6}      & 92.6      & \uline{68.6}      & \uline{79.3}      & 66.8      & \uline{78.5}    & 2.13 \\
                & \cellcolor{gray!20}Ours 
                & \cellcolor{gray!20}\textbf{81.7}  
                & \cellcolor{gray!20}\uline{79.3}  
                & \cellcolor{gray!20}\textbf{91.4}  
                & \cellcolor{gray!20}\textbf{84.4}  
                & \cellcolor{gray!20}\textbf{95.8}  
                & \cellcolor{gray!20}\textbf{75.4}  
                & \cellcolor{gray!20}\textbf{86.8}  
                & \cellcolor{gray!20}\textbf{78.2}  
                & \cellcolor{gray!20}\textbf{83.9}  
                & \cellcolor{gray!20}\uline{0.68}  \\
         \bottomrule
         \addlinespace[2mm]  
    \end{tabularx}
    \begin{tablenotes}
        \item[a] FM: foundation Model.
        \item[b] Llama: Llama3.1-8B-Instruct.
        \item[c] GPT: GPT-4o-mini.
        \item[d] For the GPT-4o-mini model, we adopted OpenAI's official pricing (\$0.15/M input tokens and \$0.6/M output tokens); for the Llama-3.1-8B-Instruct model, we referenced the pricing from OpenRouter (\$0.02/M input tokens and \$0.05/M output tokens).
    \end{tablenotes}
    
    \end{threeparttable} 
    \label{table:main_result}
\end{table*}
\begin{table*}[!h]
    \centering
    \caption{Performance Comparison Across Different Optimizers on Multiple Benchmarks}
    \renewcommand{\arraystretch}{1.4} 
    \scriptsize
    \begin{threeparttable}
    \begin{tabularx}{\textwidth}{l|XXXXXXXX|c}
        \toprule
        \multirow{2}{*}{\textbf{Optimizer}} & \multicolumn{8}{c|}{\textbf{Benchmarks}} & \multirow{2}{*}{\textbf{Avg.}} \\
        \cline{2-9}
        & \rule{0pt}{5ex}\textit{MMLU} & \textit{DROP} & \makecell{\textit{Human}\\\textit{Eval}} & \textit{MBPP} & \textit{GSM8K} & \textit{MATH} & \textit{MedQA} & \makecell{\textit{Fin}\\\textit{Eval}} & \\
        \midrule
        GPT-4o-mini         & 69.2          & 62.8          & 73.8          & 52.7          & 82.5          & 31.3          & 57.8          & 50.7          & 58.0 \\
        GPT-4o              & \textbf{79.8} & \textbf{67.4} & \uline{83.5}  & 69.8          & \textbf{92.2} & \textbf{44.6} & \textbf{65.4} & \textbf{58.5} & \textbf{68.7} \\
        Claude-3.5-sonnet   & \uline{78.7}  & 65.7          & \textbf{84.8} & \textbf{70.8} & \uline{90.3}  & \uline{43.8}  & \uline{60.3}  & 55.7          & \uline{67.1} \\
        Gemini Pro 1.5      & 76.8          & \uline{65.9}  & 81.7          & \uline{70.0}  & 89.7          & 42.8          & 58.8          & \uline{56.8}  & 66.3 \\
        DeepSeek V3         & 77.4          & \uline{64.5}  & 75.7         & 62.3  &    88.3       & 36.5          & \uline{62.2}         & 54.6  & 63.3 \\
        \bottomrule
    \end{tabularx}
    \label{table:diff_optimizer}
    \end{threeparttable}
\end{table*}
\begin{figure*}[!h]
    \begin{minipage}[t]{0.48\textwidth}
        \centering
        \includegraphics[width=\linewidth]{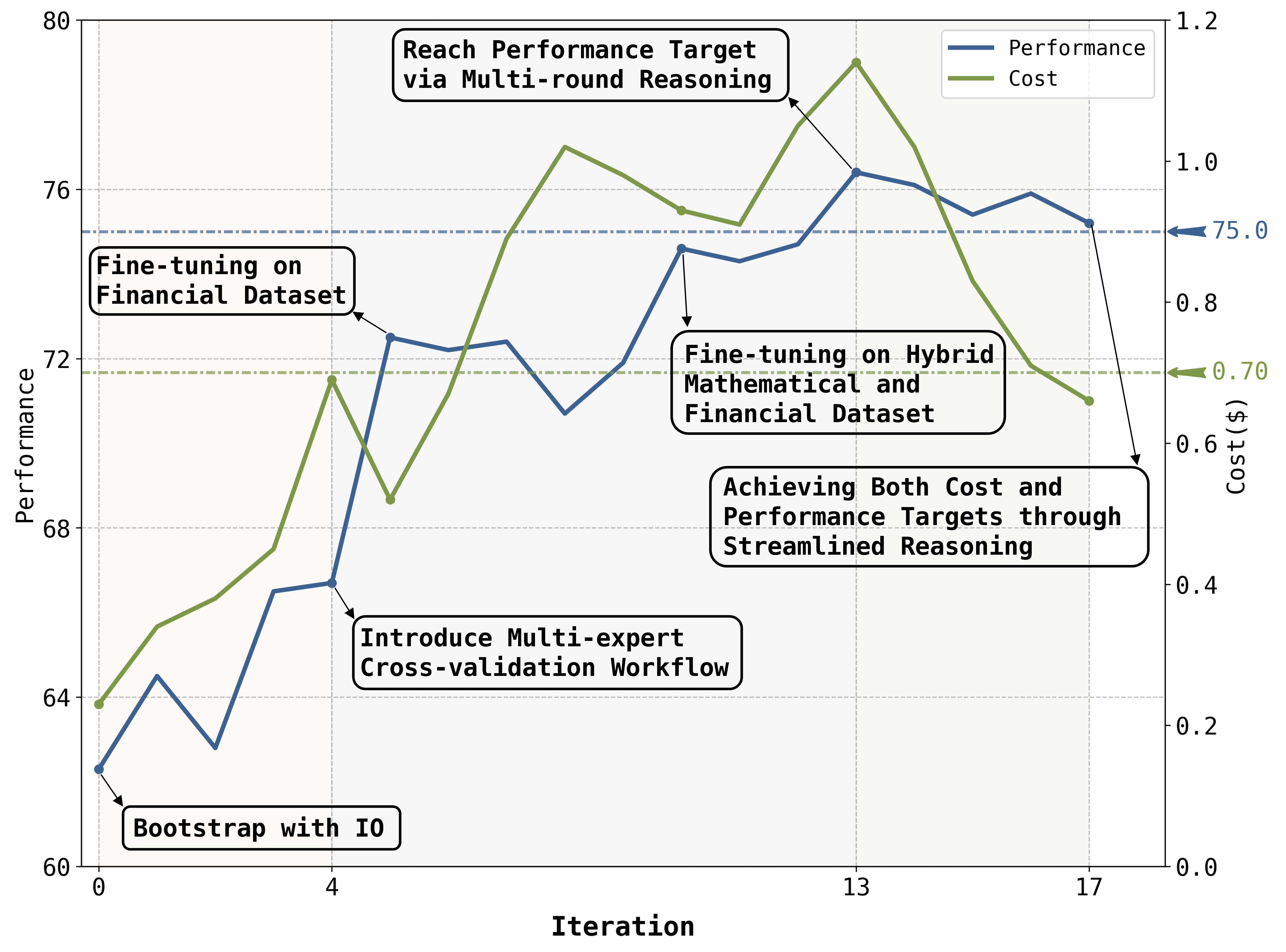}   
        \caption{The optimization process on the FinEval benchmark, where dashed lines indicate the target performance and cost}
        \label{fig:case_study}
    \end{minipage}
    \hfill
    \begin{minipage}[t]{0.48\textwidth}
        \centering
        \includegraphics[width=\linewidth]{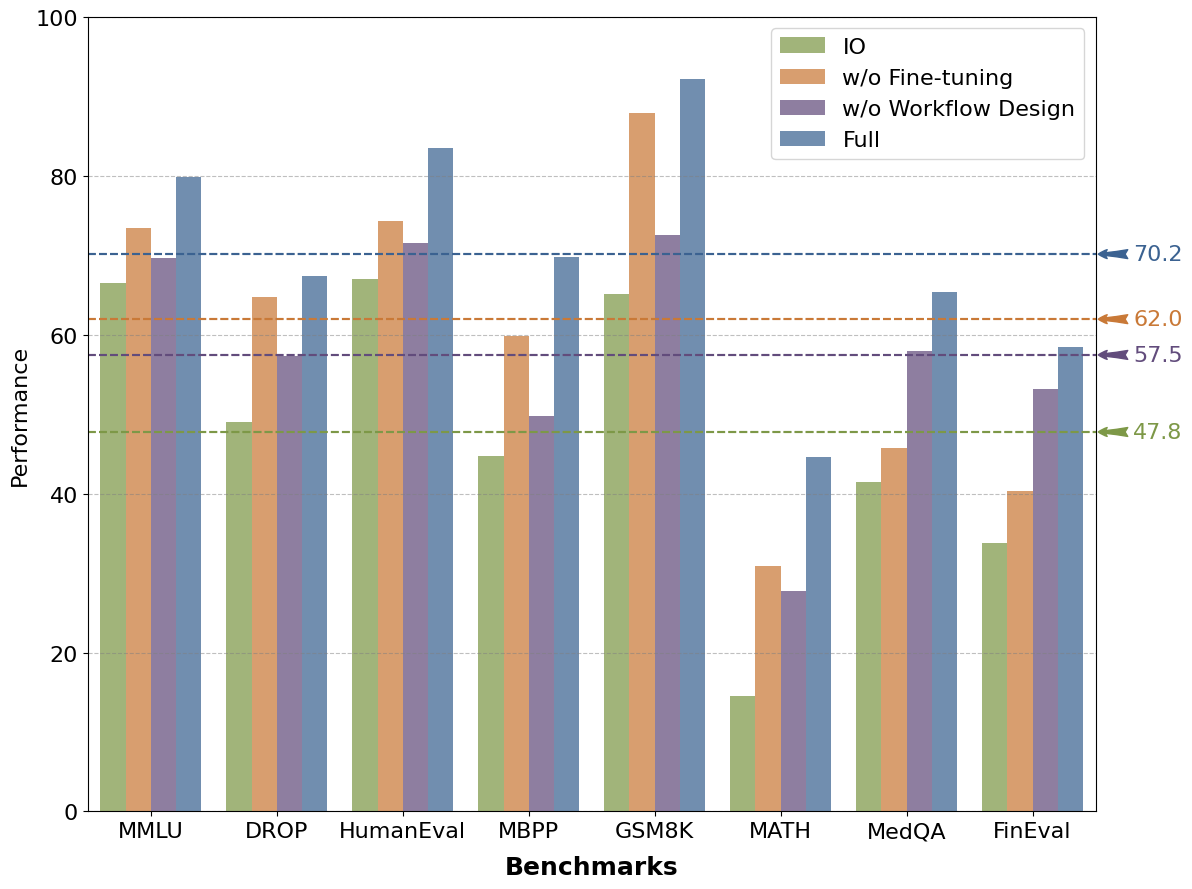} 
        \caption{Ablation study results across different benchmarks, where dashed lines indicate the average performance}
        \label{fig:ablation}
    \end{minipage}
\end{figure*}

\subsection{Main Result}
We conducted extensive experiments across eight benchmarks, with each experiment repeated three times to ensure statistical reliability. 
Table \ref{table:main_result} presents our comprehensive results, where "Avg" represents the mean performance across all benchmarks and "Cost" indicates the total inference cost. 
The results demonstrate two significant findings. 
First, AgentFactory consistently outperforms both manually designed methods and existing automated agentic system design approaches across nearly all benchmarks, with an average performance improvement of 9.1\% over current automated design methods.
This superior performance is particularly pronounced in domain-specific tasks, with notable improvements of 19.6\% in MedQA and 18.7\% in FinEval, where traditional workflow-only optimization methods showed limited effectiveness.
This performance gap can be attributed to AgentFactory's unique capability to jointly optimize both foundation models and workflow structures, enabling enhanced domain adaptation through targeted fine-tuning. 
Second, AgentFactory achieves these performance improvements while maintaining competitive operational costs, particularly when using GPT-4o-mini as the base model, reducing average costs by up to 68\%. This cost efficiency stems from our multi-objective optimization strategy, which actively balances performance requirements against computational expenses by strategically incorporating more cost-effective models when possible.

To investigate the impact of different LLM optimizers on system performance, we conducted comparative analyses across multiple powerful language models, 
as shown in Table \ref{table:diff_optimizer}. Our findings reveal two key insights: 
First, the optimization effectiveness correlates with the underlying LLM's capabilities, with more sophisticated models like GPT-4o consistently achieving better results compared to their smaller counterparts (e.g., GPT-4o-mini). 
Second, LLMs of comparable capabilities (such as GPT-4o and Claude-3.5-Sonnet) produce similar optimization outcomes, suggesting that AgentFactory's effectiveness is robust across different LLM implementations of similar capability levels. This consistency validates the framework's generalizability and independence from specific LLM models.

\subsection{Case Study: Optimizing Performance on FinEval} 
To demonstrate AgentFactory's optimization capabilities, we conducted a detailed analysis of the optimization process on the FinEval benchmark, as illustrated in Figure \ref{fig:case_study}. The experiment utilized both GPT-4o-mini and Llama3.1-8B-Instruct as available foundation models. Starting from a basic IO approach with an initial accuracy of 62.3\%, AgentFactory evolved through several key stages:
\begin{itemize}
    \item \textbf{Workflow Optimization (Iterations 1-4)}: The system discovered a Multi-expert Cross-validation Workflow, improving accuracy to 66.7\%.
    \item \textbf{Model Fine-tuning Integration (Iteration 5)}: AgentFactory introduced domain-specific fine-tuning on financial datasets, resulting in a significant performance jump to 72.5\%.
    \item \textbf{Cross-domain Enhancement (Iteration 11)}: A crucial breakthrough occurred when the optimizer identified mathematical components within FinEval tasks and strategically incorporated mathematical instruction data into the fine-tuning process, pushing accuracy to 74.6\%.
    \item \textbf{Multi-objective Optimization (Iterations 13-17)}: The system developed a Multi-Reasoning approach that exceeded the target accuracy threshold of 75\%. Subsequently, AgentFactory focused on cost optimization, successfully reducing operational expenses from \$1.14 to the target of \$0.60 through workflow refinement while maintaining performance. The optimization process terminated at iteration 17 after meeting both accuracy and cost targets.
\end{itemize}

This case study demonstrates AgentFactory's capability to automatically discover optimal agent architectures through systematic exploration, eliminating the need for manual trial-and-error in agent design.

\subsection{Ablation Study}
To validate the effectiveness of AgentFactory's optimization in the joint space of model parameters and workflow representations, we conducted ablation studies using Llama3.1-8B-Instruct as the base model, with results shown in Figure \ref{fig:ablation}. 
Direct LLM invocation (IO) serves as the baseline, while "w/o fine-tuning" and "w/o workflow design" represent optimization solely through workflow design and model fine-tuning, respectively. 
The results demonstrate that both individual components improve performance over the baseline IO approach, with the full AgentFactory pipeline surpassing the fine-tuning-only approach by 12.7\% and the workflow-design-only approach by 8.2\% on average. Notably, fine-tuning shows particularly significant improvements on domain-specific benchmarks (MedQA and FinEval), validating the importance of incorporating fine-tuning in AgentFactory's framework. 
The full AgentFactory pipeline, which combines both optimization strategies, achieves the best overall performance across all benchmarks.

\section{Conclusion}
In this paper, we presented AgentFactory, a framework that advances automated agentic system design. Prior approaches have been limited by their singular focus on workflow optimization, neglecting the potential benefits of model fine-tuning and typically optimizing for a single objective. In contrast, our work formulates automated agentic system design as an optimization problem with LLMs serving as the optimizer across a joint search space of both foundation models and workflow structures. Through our three-stage optimization pipeline, AgentFactory significantly expands the solution space while supporting multi-objective optimization across multiple dimensions. Our extensive experiments across eight benchmarks spanning five domains demonstrated significant improvements over existing methods, achieving an average performance gain of 9.1\% and particularly impressive improvements in specialized domains (19.6\% on MedQA and 18.7\% on FinEval) while maintaining lower inference costs. These results establish AgentFactory as a promising approach for developing more capable and efficient agentic systems through automated optimization.

\section*{Acknowledgments}
This work was supported by the National Natural Science Foundation of China (NSFC) under Grant
No. 92373206, the National Key Research and Development Program of Ministry of Science and
Technology under Grant No. \seqsplit{2024YFB3614200}, the Shenzhen Science and Technology Program under
Grants No. \seqsplit{JCYJ20220818100814033}, No. \seqsplit{SGDX20230116093303006}, No. \seqsplit{KJZD20231023100201003}
and No. \seqsplit{KQTD20200820113105004} and by the "Intelligent Chip" Interdisciplinary Exploration Program from the School of Electronic and Computer Engineering, Peking University.

%
%
%
\bibliographystyle{splncs04}
\bibliography{ref.bib}

\end{document}